# SsciBERT: A Pre-trained Language Model for Social Science Texts

Si Shen[1,*] Jiangfeng Liu[2] Litao Lin[2] Ying Huang[3,4] Lin Zhang[3,4] Chang Liu[2] Yutong Feng[2] Dongbo Wang[2]

*1 Group of Science and Technology Full-text Knowledge Mining, School of Economics & Management, Nanjing University of Science and Technology, Nanjing 210094, China*

*2 College of Information Management, Nanjing Agricultural University, Nanjing 210095, China*

*3 Center for Science, Technology & Education Assessment (CSTEA), Wuhan University, Wuhan, 430072, China*

*4 Center for Studies of Information Resources, School of Information Management, Wuhan University, Wuhan, 430072, China*

**\* Corresponding Email: shensi@njust.edu.cn**

**Abstract:** The academic literature of social sciences records human civilization and studies human social problems. With its large-scale growth, the ways to quickly find existing research on relevant issues have become an urgent demand for researchers. Previous studies, such as SciBERT, have shown that pre-training using domain-specific texts can improve the performance of natural language processing tasks. However, the pre-trained language model for social sciences is not available so far. In light of this, the present research proposes a pre-trained model based on the abstracts published in the Social Science Citation Index (SSCI) journals. The models, which are available on GitHub (https://github.com/S-T-Full-Text-Knowledge-Mining/SSCI-BERT), show excellent performance on discipline classification, abstract structure-function recognition, and named entity recognition tasks with the social sciences literature.

**Keywords**: Social Science; Natural Language Processing; Pre-trained Models; Text Analysis; BERT

# 1 Introduction

With the development of open access (OA), data science (DS), and natural language processing (NLP), empirical research on the social sciences that involve conventional surveys and statistical analysis methods cannot meet the growing need to analyze the contents of various academic literature, policy texts, and survey reports. Among the various social science disciplines, the use of information science methods in cross-disciplinary research has generated such overlaps as computing sociology, computing linguistics, computing law, and computing humanities (digital humanities). Meanwhile, based on the existing theories and methods of informatics, combining NLP and deep learning with the full text of academic literature to probe the relevant issues in informatics has become a hot research topic.

The data science and NLP technology methods provide a solid theoretical and technical basis for breakthroughs and innovations in social science research methods. Among them, data science focuses on combining statistics and computer science, while NLP emphasizes the combination of linguistics and computer science. To satisfy the urgent need for mass social science text analysis, NLP technology has become an indispensable part of the new generation of social science research. As a type of correspondence, the language model represents the mathematical modeling of language based on objective linguistic facts. Further, with deep learning techniques continuing to develop, pre-training techniques have been greatly improved. The method of using a vast unsupervised corpus for pre-training as a replacement for random initialization in traditional neural network models is at the core of such techniques. Language models rely on a pre-trained component to acquire a priori knowledge that supports downstream tasks for subsequent text mining. ELMo (Peters et al., 2018) and BERT (Devlin, Chang, Lee, & Toutanova, 2019) demonstrated that pre-training with a vast corpus followed by fine-tuning on downstream tasks could significantly improve the effectiveness of text mining tasks. SciBERT (Beltagy, Lo, & Cohan, 2019) also found that, compared to pre-training with a general-purpose corpus, pre-training on the domain-specific literature could improve the performance of the model on domain-specific text mining.

Based on the existing research and given the current needs of social science research, this research aims to build a pre-trained language model called SsciBERT for the intelligent processing of social science texts. The BERT-base and SciBERT were chosen as benchmark models for experiments, and the social science corpus, which was used to further pre-train these benchmarks into different variants of the SsciBERT model, was assembled. The research purpose is to find out which model performs better at NLP tasks with the social sciences literature and, with the leading model, provides a better pre-trained model to support various types of text mining and intelligent processing in social sciences texts. The main contributions of this research include the following:

(1) A social science dataset, including the abstracts and titles of SSCI papers between 1986 and 2021, is constructed from the Web of Science (WoS), which includes the abstracts and titles of SSCI papers between 1986 and 2021. The data has been cleaned, sorted, and processed to form a standard social sciences dataset.

(2) Some domain pre-trained models make due allowance for the needs of several kinds of social science research. The SsciBERT pre-trained model was built to compensate for the lack of pre-trained models in the academic domain of social science.

(3) Some statements have been verified. The constructed pre-trained model to a validation dataset was applied, and the performance to be both outstanding and reliable was verified. In addition, this also provided methodological and technical references for the construction of large-scale, high-performance automatic classification models built from and for the social sciences, as well as an abstract structure-function recognition model and named entity recognition models.

## 2 Literature review

The idea of pre-trained models was originally proposed for problems in the image field (e.g., VGG (Simonyan & Zisserman, 2014), ResNET (He, Zhang, Ren, & Sun, 2016)) and, following a good performance in this field, the technique was then applied to NLP. Pre-training uses a large-scale raw corpus to train a deep network structure, i.e., a pre-trained model with a set of model parameters. The pre-trained models can be fine-tuned directly for downstream tasks, eliminating the need to start from scratch. The pre-trained models comprise pre-trained word embeddings (PWE) and pre-trained contextual encoders (PCE) models. PWE models characterize words as low-dimensional dense distributed vectors, with representative models that include NNLW (Bengio, Ducharme, & Vincent, 2000), Word2Vec (Mikolov, Chen, Corrado, & Dean, 2013), GloVe (Pennington, Socher, & Manning, 2014), and FastText (Bojanowski, Grave, Joulin, & Mikolov, 2017). Although PWE models can capture semantic features of words, they are context-independent and cannot effectively represent higher-level features, such as word sense disambiguation and semantic roles. This is where PCE models come in, with representative models including ELMo (Peters et al., 2018), GPT (Radford, Narasimhan, Salimans, & Sutskever, 2018), and BERT. PCE differs from static word embedding. Its contextual embeddings go beyond word-level semantics, and each of its annotations is associated with the representation that acts as a function of the entire input sequence. These context-dependent representations capture the syntactic and semantic properties of multi-lingual contextual words. Since PWE models realize deep joint modeling of word meaning, grammar, and language structure, they produced fine performances in part-of-speech annotation, entity recognition, machine translation, and abstract generation. In PCE models, the ELMo dynamically adjusts the embedding of polysemous words to solve their meaning in a specific context effectively. However, this model simply splices one-way language models trained independently in the front and back directions. Hence, it has a relatively weak feature fusion capability. GPT excels at capturing longer-distance contextual information, making it more suitable for machine translation, automatic abstract, and other forward generative tasks. Nevertheless, its generalization ability is relatively weak due to the large parameter set of the model. BERT forces the model to make predictions based on the context , thus realizing deep bi-directional text representation. The emergence of BERT greatly promoted the development of pre-trained models and contributed to a series of pre-trained models based on domain-specific data, such as SciBERT (Beltagy et al., 2019), BioBERT (Lee et al., 2019), PubmedBERT (Rasmy, Xiang, Xie, Tao, & Zhi, 2021), ExpBERT (Murty, Koh, & Liang, 2020), FinBERT (Yang, Siy UY, & Huang, 2020), GreekBERT (Koutsikakis, Chalkidis, Malakasiotis, & Androutsopoulos, 2020), LEGAL-BERT (Chalkidis, Fergadiotis, Malakasiotis, Aletras, & Androutsopoulos, 2020), CovidBERT (Hebbar & Xie, 2021), SChuBERT (van Dongen, Maillette de Buy Wenniger, & Schomaker, 2020), and PDBERT (Dong, Wan, & Cao, 2021). Likewise, we managed to construct the SsciBERT model based on social sciences literature.

Given the issues we confronted when conducting this research, studies on the pre-trained model in the fields of SciBERT have been mainly reviewed. After SciBERT adds contextual information about a citation, Nicholson et al. (2021) designed a 'smart citation index' called scite, which uses SciBERT to categorize citations according to three categories: contrasting, supporting, and mentioning. Lauscher et al. (2021) took these classifications of citation intent and prepared a new multi-sentence multi-intent framework to analyze the context of one specific citation. To predict clusters of mentions on cross-document coreference resolution, Cattan et al. (2021) conducted experiments with SciBERT. Viswanathan, Neubig, and Liu (2021) extracted citation graph information by designing the architecture of the model integrated with SciBERT. Reviewing deep learning for citation function classification, Medić and Šnajder (2020) focused on the training process and application of SciBERT. To support cite-worthiness detection, Wright and Augenstein (2021) fine-tuned SciBERT with the CITEWORTH dataset. Based on the PUBHEALTH dataset, Kotonya and Toni (2020) fine-tuned models by SciBERT to solve the domain task of fact-checking label predictions. Using a corpus of abstracts from science, technology, and medicine, Brack, D'Souza, Hoppe, Auer, and Ewerth (2020) trained a domain-independent classifier with SciBERT. Kuniyoshi, Makino, Ozawa, and Miwa (2020) created a corpus of 243 papers on the synthesis process for all-solid-state batteries, finding that SciBERT achieved the best score on OPERATION. By employing a new PUBHEALTH dataset containing 11.8K health insurance claims, Kotonya and Toni (2020) compared SciBERT to other models on a fact-checking task and gained the best prediction results with SciBERT. By designing an annotation scheme for the materials science domain, Friedrich et al. (2020) constructed a corpus called SOFC-Exp and identified the experiment describing sentences with SciBERT. D'Souza et al. (2020) created a dataset called Science, Technology, Engineering, and Medicine for Scientific Entity Extraction, Classification and Resolution (STEM-ECR) and put it into the use of training a domain-independent scientific entity extraction system based on SciBERT. Asada, Miwa, and Sasaki (2020) utilized SciBERT to obtain the drug description representation of the target drugs as a prerequisite work to extract drug-drug interactions from the literature. For the NLP task of understanding natural premise selection, Ferreira and Freitas (2020) found that BERT trained from the natural language-premise selection (NL-PS) corpus had even better performances than BERT. GEANet-SciBERT is a mixture made up of domain knowledge and SciBERT, which achieves an absolute improvement in experiments (Huang, Yang, & Peng, 2020). In the review of opportunities and challenges of text mining in the science of materials research, Kononova et al. (2021) emphasized the importance of SciBERT relating to future academic text mining. A SciBERT-LSTM was used in the SemEval-2021 Task 11, posting the best score of 37.83% (D'Souza, Auer, & Pedersen, 2021). Park and Caragea (2020) trained a model for scientific keyphrase identification and classification (SKIC) based on SciBERT and BERT and found that SciBERT produced a better performance than BERT.

With the development of pre-trained techniques, pre-trained models are becoming more and more important as basic computational support resources for text mining and information retrieval in natural language processing. And one important direction of the development of a pre-trained model is to combine with domain-specific data and complete the construction of a domain-specific pre-trained model through fine-tuning techniques. From the application reviews of the SciBERT model, this model has been widely used and promoted in the field of scientific and academic literature, which plays a role in domain knowledge support for both informatrics of academic texts and knowledge extraction of academic texts. On the basis of the existing pre-trained techniques, this research attempts to explore the construction of the SsciBERT model based on the academic texts of social science by summarizing the specific application of the SciBERT model.

## 3 Data and methodology

### 3.1 Dataset

Building a language model from the features of large-scale text datasets is a basic tool for implementing pre-training tasks in NLP. The size and quality of the corpus used for the pre-training phase have a decisive impact on the final performance of the language model. Therefore, abstracts from the WoS core set were chosen as the training data. The abstract is a highly condensed summary of the main content of one research paper and gives an overall description of the scientific research process. First, abstracts usually provide rich semantic knowledge. This is because abstracts have relatively prominent structural features in that each sentence of the abstract performs a certain function of summarizing the key points of various aspects of the whole study. The most famous structural division of abstracts proposed by Graetz (1982) and Swales (1990), is called IMRD-model. IMRD divides abstracts into Introduction, Method, Result, and Conclusion according to the content of research papers. This criterion for abstract division corresponds to the most widely known 'quadrinomial' text structure of scientific and technical papers (IMRaD) (Sollaci & Pereira, 2004), and serves as a common writing model for academic articles. Overall, today's academic papers follow this writing structure, and papers from core journals further ensure this rigor. Second, the abstract has high semantic integrity because it is in plain text format, differing from the content in the main body, that needs to be combined with multimodal information such as charts to accurately understand the meaning of the sentence. Third, abstracts are easier to obtain when comparing with the full text of papers. Thus the abstract data from the core journals of social sciences are selected to ensure the quality and structural consistency of the training corpus.

In the process of corpus construction, the bibliographic data of the SSCI papers published between 1986 and 2021 from WOS with ISSN as the retrieval approach was downloaded. As for data cleaning, we removed duplicate entries and entries with missing abstracts. Blank lines and special characters were also removed, and each line of the resulting corpus was the abstract from one academic literature. The final dataset consists of 2,964,743 abstracts from 3,250 journals. Table 1 and Fig. 1 show the basic information of the dataset.

**Table 1 Basic statistical information of social science abstract data**

| Basic information | All abstract statistical information |
|---|---|
| Total number of abstracts | 2,964,743 |
| Total number of words in the abstracts | 503,910,614 |
| Cumulative number of non-repeated words in the abstracts | 4,488,767 |
| The average number of total words per abstract | 169.96 |

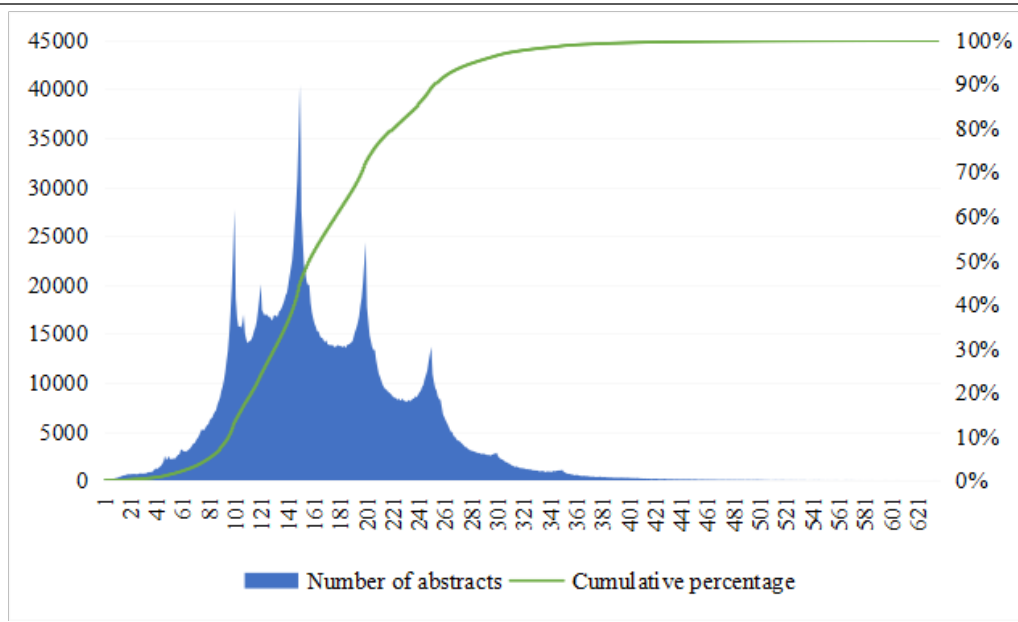

**Fig. 1 Lengths of the SSCI abstracts**

The statistics show the lengths of the abstracts and that more than 95% of them ranged from 50 to 300 words. However, it is worth noting that the sentence segmentation function of the BERT tokenizer increases the sequence length of English texts to some extent. Nevertheless, the length of most segmented sentences were still within 512 units long without being forcibly truncated, thus preserving coherent semantic features, which contributes to the overall performance of the constructed SsciBERT model.

**Table 2 The distribution of social science abstract data in accordance with discipline**

| WoS Categories | Total | with % | WoS Categories | Total | with % |
|---|---|---|---|---|---|
| Economics | 360428 | 5.75% | History | 98851 | 1.58% |
| Public, Environmental & Occupational Health | 338661 | 5.40% | Social Sciences, Interdisciplinary | 98679 | 1.57% |
| Psychiatry | 323004 | 5.15% | Linguistics | 97847 | 1.56% |
| Psychology, Multidisciplinary | 200850 | 3.20% | Environmental Sciences | 97082 | 1.55% |
| Education & Educational Research | 181963 | 2.90% | Psychology, Developmental | 90765 | 1.45% |
| Nursing | 170079 | 2.71% | Anthropology | 88231 | 1.41% |
| Health Policy & Services | 162820 | 2.60% | International Relations | 86703 | 1.38% |
| Political Science | 162504 | 2.59% | Law | 84586 | 1.35% |
| Management | 158535 | 2.53% | Business, Finance | 79348 | 1.27% |
| Environmental Studies | 155686 | 2.48% | Geography | 78430 | 1.25% |
| Health Care Sciences & Services | 140394 | 2.24% | Information Science & Library Science | 76515 | 1.22% |
| Psychology, Clinical | 134648 | 2.15% | Rehabilitation | 74813 | 1.19% |
| Psychology, Experimental | 129307 | 2.06% | Substance Abuse | 74064 | 1.18% |
| Gerontology | 126946 | 2.03% | Psychology, Applied | 70574 | 1.13% |
| Psychology | 126462 | 2.02% | Area Studies | 68077 | 1.09% |
| Business | 123600 | 1.97% | Social Sciences, Biomedical | 67338 | 1.07% |
| Sociology | 122352 | 1.95% | Communication | 66985 | 1.07% |

To better exhibit the disciplinary characteristics of the training dataset, the discipline distribution of the SSCI abstract dataset that contains 152 WoS Categories was counted. Due to the excessive number of disciplines, Table 2 shows all the disciplinary classifications (34 types) with more than 1% and their occurrences in the dataset in descending order. In addition, there is some interdisciplinary literature in the SSCI dataset, and the discipline of the interdisciplinary was counted as one count in the statistics. As can be seen from Table 2, the discipline with the highest percentage of the SSCI abstract dataset is Economics, accounting for about 5.75%, while that Communication accounts for about 1.07%, with the lowest percentage. Public, Environmental & Occupational Health and Psychiatry are between the two. Among the data set, only the top three disciplines account for more than 5%. Furthermore, Political Science, Management, Psychology, Business, Sociology, History, and other traditional disciplines in social science are represented in Table 2, and their proportion fluctuates no more than 2%, indicating a more even distribution of disciplines in terms of the summary dataset. The training text and the testing text were then divided by a ratio of 99:1 to form two datasets (Muraina, 2022). To split the dataset in this way, the reason is that expanding the proportion of training data as much as possible can ensure the full fit of the model. At the same time, due to the large scale of full data, sufficient samples can be guaranteed to obtain relatively objective evaluation results when calculating the perplexity, even though the data of the validation set only accounts for 1% of the total data volume.

## 3.2 Establishment of pre-trained models

The proposed third paradigm of NLP has further revolutionized this research methodology (Viswanathan et al., 2021). The basic paradigm of using a large-scale unlabeled corpus to pre-train a model followed by fine-tuning with a small-scale dataset of labeled text has almost become the preferred choice for processing text data. Fig. 2 shows the process of this research.

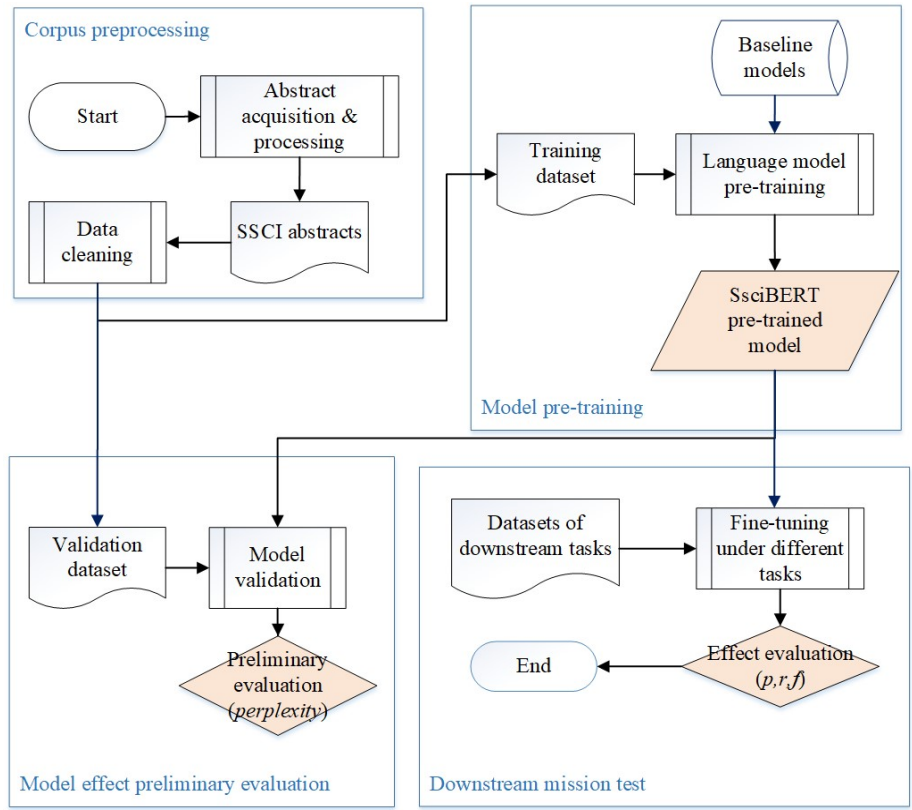

**Fig. 2 Process of the construction of the pre-trained model and the verification of its performance**

The training method for a large-scale language model is divided into training from scratch and continuous pre-training. The training from scratch method requires building a new vocabulary based on the text and retraining the parameters of each layer of the model, which usually demands a large-scale corpus to avoid overfitting problems. By contrast, pre-training based on an existing pre-trained model means further optimization of some already-trained parameters. Thus, it only requires a relatively small amount of data. In some ways, we refer to a corpus of around 500 million words, and a much larger dataset containing billions of words is necessary for training from scratch.

Furthermore, the type of corpus used is academic texts written in English, which resembles the originally trained corpus of BERT or SciBERT. Considering these factors, unlike previous studies that adapt training from scratch strategy to build a whole new pre-trained language model, we opted to do further pre-training on the already pre-trained BERT and SciBERT models with a corpus of unlabeled social science abstracts. Meanwhile, previous studies have also shown that continuous training can obtain equivalent or even better domain-specific models than training from scratch.

BERT is a pre-trained language model for text processing released by Google's AI team in 2018. BERT's main module is a bidirectional Transformer encoder (Vaswani et al., 2017) that learns the semantic representation of an input sequence. BERT's pre-training comprises two unsupervised tasks: a mask language model (MLM) pre-training objective and a next sequence prediction (NSP) task. BERT achieved state-of-the-art results at that time on 11 NLP tasks. The second benchmark model is based on SciBERT, a model pre-trained for scientific paper processing built by the Allen Institute for Artificial Intelligence. This model grapples with academic literature in the fields of computing and biology for pre-training. The models' performances were tested by researchers with several typical NLP tasks, and the verification result surpassed that of the BERT-base model.

In this research, the case-sensitive BERT-base-cased model and the SciBERT-scivocab-cased model are chosen as our two benchmark models for further pre-training on social sciences abstracts. SciBERT is selected as one of the baseline models because it is specially pre-trained based on scientific and academic text corpus, which has a relatively excellent effect on academic literature processing and conforms to the application background of bibliometrics research. The BERT-base-cased model is used for the comparison experiment, of which the purpose is to testify whether the model parameters trained by the natural science literature corpus can fit the social science paper data set more efficiently than the model trained by the general corpus. After the SsciBERT models were trained, their performance was preliminarily evaluated by perplexity. Finally, specific tasks such as discipline title and abstract classification, abstract sentence classification, and name entity recognition were employed to further validate the performance of the model.

## 4 Training and evaluating model

### 4.1 Pre-training parameters and environment configuration

In the corpus for pre-training, each line is the title or abstract of an article. Statistically, 99% of the lines have less than 512 words. Therefore, in the experiments, the maximum sequence length of 512 is set. In addition, the line-by-line parameter is also set to process each line individually instead of mixing all lines and then performing an even-length cut. The initial learning rate is set to 5e-5. The transformers (Wolf et al., 2019) package's warm-up function allows the learning rate to increase rapidly to the initial learning rate during the pre-training of the model; then, it decreases gradually to 0. This mode allows the model to stabilize rapidly during the initial training phase and then converge faster with subsequent training. A gradually decreasing learning rate also considers the need for both converging rate and training accuracy. In a comparison pre-experiment, it is found that when the learning rate is relatively low, the accuracy is higher. Accordingly, the learning rate is set to an initial 2e-5 for subsequent experiments. Based on the server configuration and the comparison test, the training batch size is set to 32. In terms of the common experimental practice, the data is trained and the effects for multiple rounds are compared. From the SciBERT literature, it is found that two- and four-round training produces the desired results. Table 3 lists the specific parameters.

Table 3 Pre-training hyperparameter

| Hyperparameter | SSCI-BERT-e2 SSCI-SciBERT-e2 | SSCI-BERT-e4 SSCI-SciBERT-e4 |
|---|---|---|
| max_seq_length | 512 | 512 |
| learning_rate | 2e-05 | 2e-05 |
| train_batch_size | 64 | 64 |
| eval_batch_size | 64 | 64 |
| num train epochs | 2 | 4 |
| line_by_line | True | True |

The experiments are performed on two Quadro RTX 8000 GPUs with 45GB of graphic memory for parallel pre-training on a Linux server (Ubuntu 20.04.3). The code used is the mask language modeling sample code provided by Transformers. It costed the four epochs of pre-training based on SciBERT 62.4 hours in total, and the four epochs of pre-training based on BERT-base 63.2 hours in total.

## 4.2 Evaluation of pre-training model

The direct method of determining the quality of a language model is to apply the language model to a specific NLP task and then observe the difference between the model and other models with regard to that specific task. However, perplexity provides a new way of solving this problem (Chen,et al.,1998). Regarding large differences in perplexity, the lower the perplexity, the better the fit of the pre-trained model to the real sentence and the better of the model's performance. As to the perplexity calculation, suppose there exists a sentence $s = (w_1, w_2, w_3, ..., w_{n-1}, w_n)$, $w_n$ represents the NTH word in the sentence, and then the probability of the whole sentence can be expressed as $P = (w_1 w_2 w_3 ... w_{n-1} w_n)$, with which the operation can be performed. The perplexity is defined as the probability of the sentence to the power -1/n. The related formula is as follows:

$$PP(W) = P(w_1 w_2 ... w_N)^{-\frac{1}{N}} = \sqrt[N]{\frac{1}{P(w_1 w_2 ... w_N)}} \quad (1)$$

To ensure the rigor of the experimental results, several sets of pre-training experiments were conducted by adjusting the number of abstracts and the number of training rounds. The perplexity of the pre-trained language model is shown in the table below.

Table 4 Perplexity of pre-trained models

| Model | Data | Benchmark Model | Perplexity |
|---|---|---|---|
| Bert-base-cased | -- | -- | 13.313 |
| Scibert-scivocab-cased | -- | -- | 9.331 |
| SSCI-BERT-e2 | 1986-2021 | Bert-base-cased | 5.754 |
| SSCI-SciBERT-e2 | 1986-2021 | Scibert-scivocab-cased | 5.330 |
| SSCI-BERT-e4 | 1986-2021 | Bert-base-cased | 5.583 |
| SSCI-SciBERT-e4 | 1986-2021 | Scibert-scivocab-cased | 5.195 |

As shown in Table 4, the perplexity of pre-trained models based on abstracts and titles of articles in social sciences is relatively low. Given that the test dataset consists of titles and abstracts of the journal articles, the testing dataset's contents can be considered normal and comprehensible sentences. Therefore, the performance of the language model with lower perplexity in the specific task may surpass that of the model with higher perplexity. These models on specific text mining tasks are subsequently further verified.

### 4.3 Model verification

Three NLP tasks are selected to verify the adequate performance of the proposed four pre-trained models based on the SSCI dataset. The first one was a classification task with Journal Citation Reports (JCR) social science disciplines, the second was to identify structures in the SSCI abstracts, and the third one was to recognize the software entities in the full text of *Scientometrics*. The use of a high-quality standard dataset is a prerequisite to the reliability of the model validation. Considering that the constructed model is a pre-trained model for social sciences, the relevant data from SSCI papers were chosen as data sources. BERT and SciBERT were used as the benchmark models for performance comparison.

### 4.3.1 Verification task description

**(1) Classification task for JCR social science disciplines**

Today, the number of research fields is constantly expanding, along with which the number of academic papers has increased dramatically. In this context, automatically classifying academic papers according to disciplines could greatly benefit the indexing of literature and the construction of knowledge bases. Such classification research could save a researcher exploration time and improve their efficiency in finding the required content in a vast sea of literature.

Hence, to verify the effectiveness of the pre-trained models on the classification task, JCR disciplines classification was performed based on the titles and abstracts of the SSCI journal papers. In this task, journal papers published between 2006 and 2020 were assembled. Discipline matching against the JCR discipline list was then conducted according to the ISSN numbers of the journals. After the dataset was cleaned, 500 pieces of data were extracted from each discipline as the dataset of the classification task for JCR social science disciplines. A total of 23,000 pieces of title data and 22,000 pieces of abstract data from 46 disciplines were obtained. Three types of data sets are used in this task for discipline classification to verify the performance of the model. They are pure title, pure abstract, and title and abstract mixed dataset. The discipline categories are shown in Table 5.

**Table 5 Classification of JCR social science disciplines**

| NO. | Category | NO. | Category |
| --- | --- | --- | --- |
| 1 | Anthropology | 24 | Industrial Relations & Labor |
| 2 | Area Studies | 25 | Information Science & Library Science |
| 3 | Business | 26 | International Relations |
| 4 | Business, Finance | 27 | Law |
| 5 | Cultural Studies | 28 | Linguistics |
| 6 | Communication | 29 | Management |
| 7 | Criminology & Penology | 30 | Nursing |
| 8 | Demography | 31 | Political Science |
| 9 | Development Studies | 32 | Psychology, Multidisciplinary |
| 10 | Economics | 33 | Public Administration |
| 11 | Education & Educational Research | 34 | Public, Environmental & Occupational Health |
| 12 | Education, Special | 35 | Regional & Urban Planning |
| 13 | Environmental Studies | 36 | Rehabilitation |
| 14 | Ethics | 37 | Social Issues |
| 15 | Ethnic Studies | 38 | Social Sciences, Biomedical |
| 16 | Family Studies | 39 | Social Sciences, Interdisciplinary |
| 17 | Geography | 40 | Social Sciences, Mathematical Methods |
| 18 | Gerontology | 41 | Social Work |
| 19 | Health Policy & Services | 42 | Sociology |
| 20 | History | 43 | Substance Abuse |
| 21 | History & Philosophy of Science | 44 | Transportation |
| 22 | History Of Social Sciences | 45 | Urban Studies |
| 23 | Hospitality, Leisure, Sport & Tourism | 46 | Women's Studies |

### (2) Task for Identifying Abstract Structures

The abstract is a concise description of the content of literature. However, due to the fact that most abstracts lack a unified structure or markup, how to identify different functional units in abstracts has become an important research area for knowledge mining academic literature. In this task, the abstracts of SSCI journal papers published between 2008 and 2020 were first extracted from WOS. Then, the original IMRD paradigm was improved by subdividing the "problem" structure into two parts: Background and Purpose. The reason for adopting this strategy is that the background and purpose show high semantic similarities. In this way, the introduction is divided into Background and Purpose. On the one hand, it gives rise to more difficulties for the verification task so as to fully test the performance of the model more; on the other hand, it also makes the corpus more accurate and can better serve for future research. Then, the functional structure of the abstracts was annotated sentence by sentence according to the five categories of Background, Purpose, Methods, Results, and Conclusions (BPMRC). Finally, a total of 1,378,276 structural annotations were obtained from this process. Given the large size, the data were divided into testing and training datasets in a ratio of 1:9.

**(3) Task for Software Entities Recognition in *Scientometrics***

In the era of big data, the software is already playing a significant role in academic research. Identifying software entities provides new insights into how we understand academic research development. In order to build a high-quality software entity dataset, we have done the following series of work: First, full-text data in HTML format published in *Scientometrics* between 2010 and 2020 was first acquired by using a homemade web crawler, and then we parsed the data into plain text. Next, the full-text was split into sentences of character length up to 512 by the StanfordCoreNLP toolkit. Then, we deployed the BRAT annotation platform and uploaded the full text into it in sentence units. We developed a detailed annotation specification, and then convened ten current graduate students in informatics to manually annotate the software entities in the sentences. After the annotation was completed, the annotation results were checked and proofread to ensure the quality of the annotation. The kappa coefficient was used as the consistency test of annotation, and the consistency coefficient of software entity annotation is equal to 0.821. Finally, a total number of 13,269 software entities were identified. During the process of annotation, BMES（B-Software, M-Software, E-Software, and S-Software）tagging scheme was used. If the software entity consists of only one word, it will be labeled as S-Software; characters that do not represent software entities will be labeled as O. Finally, the ratio between testing and training datasets is 1:9. The training sentence "Two other computer programs are Citespace and Network Workbench Tool." is annotated as follows:

```
Two            O
other          O
computer       O
programs       O
are            O
Citespace      S-software
and            O
Network        B-software
Workbench      M-software
Tool           E-software
.              O
```

**Fig. 3 Example annotation of a software entity dataset**

### 4.3.2 Benchmark model and verification index

To verify the performance of the four pre-trained models in various NLP tasks, two benchmarks – BERT-base-cased and SciBERT-scivocab-cased are selected for comparison. Meanwhile, six indicators were used to evaluate the experimental performance, i.e., Accuracy, Precision, Recall, F1-score, Macro-average, and Weighted average. The formulations of each indicator are as follows:

Table 6 shows the confusion matrix for the classification quality measure of the label-based evaluation. The four possible classification results are: True Positive (TP), False Positive (FP), True Negative (TN), and False Negative (FN). Confusion matrix has long been present in the evaluation of scientific models and engineering applications, and it can reveal the classification result and other characteristics of the model more clearly. Based on confusion matrix, four evaluation metrics, i.e., Accuracy, Precision, Recall, and F1-Score are given in equations (2)-(5). Among these metrics, accuracy represents the rate of correct prediction in all samples; precision is the degree of accuracy in predicting positive sample results; recall is the degree of comprehensiveness in predicting positive sample results; $F1$ value is the average harmonic value of accuracy and recall. When the categories of the experimental samples are heterogeneous, the model's effect would not be verified by the accuracy rate itself. And thus, multiple averaging methods are employed to reconcile the effects brought by the data, and the macro-average $F1$ value and the weighted average $F1$ value are used for the evaluation of model performance.

**Table 6 Confusion matrix**

| Real category | Prediction Category | |
|---|---|---|
| | Positive example | Negative Example |
| Positive example | TP | FN |
| Negative Example | FP | TN |

$$Accuracy = \frac{TP+TN}{TP+TN+FP+FN} \quad (2)$$

$$Precision = \frac{TP}{TP+FP} \quad (3)$$

$$Recall = \frac{TP}{TP+FN} \quad (4)$$

$$F1-score = \frac{2 precision * recall}{precision * recall} \quad (5)$$

Macro-average is the arithmetic average of the value of each statistical indicator for all categories, with specific indicators such as macro-precision, macro-recall, and macro-$F1$-score. The specific formulas for the indicator are as follows:

$$macro-precision = \frac{1}{n}\sum_{i=1}^{n} precision_i \quad (6)$$

$$macro-recall = \frac{1}{n}\sum_{i=1}^{n} recall_i \quad (7)$$

$$macro-F1score = \frac{2 precision_{macro} * recall_{macro}}{precision_{macro} + recall_{macro}} \quad (8)$$

The weighted average uses the weight of the proportion of the number of samples in each category to the total number of samples in all categories and then calculates the average. The specific indicators include weighted precision, weighted recall, and weighted $F1$-score. The specific formulas for the indicators are as follows:

$$weighted - precision = \sum_{i=1}^{n} precision_i * f_i \quad (9)$$

$$weighted - recall = \sum_{i=1}^{n} recall_i * f_i \quad (10)$$

$$weighted - F1score = \frac{2 precision_{weighted} * recall_{weighted}}{precision_{weighted} + recall_{weighted}} \quad (11)$$

### 4.3.3 Verification results analysis

**(1) Disciplines classification results based on different pre-trained models**

As mentioned, the classification performance of the SSCI pre-trained models was verified based on three corpora with a classification task involving the JCR discipline system. The model parameters are given in Table 7.

**Table 7 Task model parameters of JCR social science discipline classification**

| Parameters | Value |
|---|---|
| max_seq_length | 512 |
| train_batch_size | 32 |
| gradient_accumulation_steps | 4 |
| eval_batch_size | 128 |
| learning_rate | 2e-5 |
| adam_epsilon | 1e-6 |

Experimental results are shown in Table 8. SSCI-SciBERT-e2 based on SciBERT achieved the best results in the aspects of accuracy, $F$1-score under macro-average, as well as $F$1-score under weighted average. It obtained a weighted average $F$1 of 37.25%, which is an improvement of 3.33% compared to SciBERT. The comparison revealed that the performance of the two models based on SciBERT in this experiment surpassed that of the two models based on BERT. Although the proposed four models showed a significant improvement in performance compared to their respective original models, the weighted average $F$1 score was less than 40%.

**Table 8 Discipline classification results of JCR social science titles ($F$1-score)**

| Model | accuracy | macro avg | weighted avg |
|---|---|---|---|
| Bert-base-cased | 28.43% | 22.06% | 21.86% |
| Scibert-scivocab-cased | 38.48% | 33.89% | 33.92% |
| SSCI-BERT-e2 | 40.43% | 35.37% | 35.33% |
| SSCI-SciBERT-e2 | 41.35% | 37.27% | 37.25% |
| SSCI-BERT-e4 | 40.65% | 35.49% | 35.40% |
| SSCI-SciBERT-e4 | 41.13% | 36.96% | 36.94% |

As shown in Table 9, the SSCI-SciBERT-e2 obtained the highest values in the classification task with respect to the accuracy, macro-average, and weighted average $F$1-score. Among all the models, it produced the best results. The other three pre-trained models also outperformed the two benchmark models in terms of classification performance. Compared with classification results obtained based on the title corpus, the overall results of classification tasks using the abstract corpus were better, with the highest weighted $F$1-score of 57.12%. This is probably attributed to the fact that the unit data length of the abstract corpus is longer. As a result, the model can acquire more accurate features by encoding long text, thus improving the text classification results.

**Table 9 Discipline classification results of JCR social science abstracts ($F$1-score)**

| Model | accuracy | macro avg | weighted avg |
| --- | --- | --- | --- |
| Bert-base-cased | 48.59% | 42.80% | 42.82% |
| Scibert-scivocab-cased | 55.59% | 51.40% | 51.81% |
| SSCI-BERT-e2 | 58.05% | 53.31% | 53.73% |
| SSCI-SciBERT-e2 | 59.95% | 56.51% | 57.12% |
| SSCI-BERT-e4 | 59.00% | 54.97% | 55.59% |
| SSCI-SciBERT-e4 | 60.00% | 56.38% | 56.90% |

Table 10 shows the results for the classification task with both the abstract and title data. The SSCI-SciBERT-e4 obtained the highest values in terms of accuracy, macro-average, and weighted average $F$1-score. With a weighted average of 60.75%, it obtained the best results among all the models. The other three pre-trained models all surpassed the benchmark models. Additionally, the four-round pre-trained models outperformed the two-round ones. As the size of the corpus increased, the overall performance of the optimal pre-trained model using titles and abstracts was 23.5% and 3.63% higher than the title and abstract, respectively.

**Table 10 Discipline classification results of JCR social science titles and abstracts ($F$1-score)**

| Model | accuracy | macro avg | weighted avg |
| --- | --- | --- | --- |
| Bert-base-cased | 58.24% | 57.27% | 57.25% |
| Scibert-scivocab-cased | 59.58% | 58.65% | 58.68% |
| SSCI-BERT-e2 | 60.89% | 60.24% | 60.30% |
| SSCI-SciBERT-e2 | 60.96% | 60.54% | 60.51% |
| SSCI-BERT-e4 | 61.00% | 60.48% | 60.43% |
| SSCI-SciBERT-e4 | 61.24% | 60.71% | 60.75% |

**(2) Identification results of abstract structures**

The classification and recognition of abstracts is a classic issue in applied linguistics research. From an NLP perspective, the problem lies in text classification. Although a host of scholars have constructed abstract recognition models based on rules and machine learning methods, their effectiveness needs to be further stressed. The pre-trained language model has largely improved the effectiveness of acquiring the potential semantic knowledge of academic texts. Nevertheless, few studies have attempted to improve the pre-trained model for abstract recognition. In this experiment, the label set <B, P, M, R, C> corresponds to each of the structures in the BPMRC paradigm. The sentences belonging to different abstract structures were inputted into the four pre-trained models, and the four pre-trained models were tested with regard to their performance in assigning them to the correct category. P (Precision), R (Recall), and $F$1-score were used as indicators to evaluate performance. The experimental parameters are the same as those shown in the previous section for the discipline classification experiment. Table 11 shows the results.

**Table 11 Classification results of SSCI-abstract structural function identification ($F$1-score)**

|   | Bert-base-cased | SSCI-BERT-e2 | SSCI-BERT-e4 | support |
|---|---|---|---|---|
| B | 63.77% | 64.29% | 64.63% | 224 |
| P | 53.66% | 57.14% | 57.99% | 95 |
| M | 87.63% | 88.43% | 89.06% | 323 |
| R | 86.81% | 88.28% | 88.47% | 419 |
| C | 78.32% | 79.82% | 78.95% | 316 |
| accuracy | 79.59% | 80.90% | 80.97% | 1377 |
| macro avg | 74.04% | 75.59% | 75.82% | 1377 |
| weighted avg | 79.02% | 80.32% | 80.44% | 1377 |
|   | Scibert-scivocab-cased | SSCI-SciBERT-e2 | SSCI-SciBERT-e4 | support |
| B | 69.98% | 70.95% | 70.95% | 224 |
| P | 58.89% | 60.12% | 58.96% | 95 |
| M | 89.37% | 90.12% | 88.11% | 323 |
| R | 87.66% | 88.07% | 87.44% | 419 |
| C | 80.70% | 82.61% | 82.94% | 316 |
| accuracy | 81.63% | 82.72% | 82.06% | 1377 |
| macro avg | 77.32% | 78.37% | 77.68% | 1377 |
| weighted avg | 81.60% | 82.58% | 81.92% | 1377 |

The experiments on recognizing abstract structure-function show that the $F$1-score of SSCI-SciBERT-e2 weighted average reached 82.58%. In addition, its $F$1-score of accuracy and macro-average were also the highest among all models, thus achieving the best overall performance. Following next is SSCI-SciBERT-e4, with the $F$1-score of the weighted average of 81.92%. All else being equal, the average recognition result of the SciBERT-based pre-trained models generally appeared to be superior to the models trained based on BERT. Nevertheless, the effect of different corpus and pre-training epochs on the model is indiscernible. Specifically, SSCI-SciBERT-e2 achieved the best results, with an average of 90.12%. The excellent performance of SSCI-SciBERT in abstract structure classification shows that the continuous training based on social science abstract texts can fairly support the model in extracting deep syntactic and semantic features of social science texts. The above experimental results reveal that the proposed models of SSCI-SciBERT-e2 and SSCI-BERT-e4 possess strong social science characteristics, both of which enjoy considerable advantages in terms of NLP tasks for social science texts, with SSCI-SciBERT-e2 being more suitable for the intelligent processing of academic literature in the social sciences.

**(3) Recognization results of software entities in *Scientometrics***

Named entity recognition is a common experiment to evaluate model performance. In our experiment, the performance of models is examined through the recognition performance of software entities in academic papers. P (precision), R (recall) and $F$1 score are used for assessing performance of models. Table 12 shows the relevant experimental results.

**Table 12 Recognize results of software in *Scientometrics***

| Model | P | R | F1 |
|---|---|---|---|
| Bert-base-cased | 79.19% | 84.52% | 81.77% |
| Scibert-scivocab-cased | 79.73% | 85.49% | 82.51% |
| SSCI-BERT-e2 | 80.94% | 85.81% | 83.30% |
| SSCI-SciBERT-e2 | 80.46% | 84.20% | 82.29% |
| SSCI-BERT-e4 | 82.71% | 83.24% | 82.97% |
| SSCI-SciBERT-e4 | 81.43% | 83.00% | 82.21% |

Table 12 shows the metrics of the six models in terms of software named entity recognition. Among the six models, the SSCI-ABS-BERT-e2 model achieves the highest value for $F$1 score and recall while the SSCI-ABS-BERT-e4 model for precision. The models continuously pretrained on BERT perform better than BERT in terms of $F$1 score. With the increasing training epochs, higher precision will be associated with the lower recall, which may indicate that overfitting has occurred. Overall, the model proposed in the present research performs better regarding software entity recognition. To the best of our knowledge, the model named SSCI-BERT-e2 is the best model we recommended for the Named Entity Recognition task.

## 5 Discussion

From the perspective of perplexity, all models proposed in this paper have a perplexity of 5 to 6 on the test dataset, while the benchmark model without pre-training around ten on the test dataset, which confirms that the pre-training experiment made the models fit the abstract data of the academic papers well. By comparing the perplexity of several models, it is found that the models with four epochs of training had a lower perplexity than those with two epochs of training. The model with the lowest perplexity was SSCI-SciBERT-e4, with SciBERT as the benchmark model with four epochs of pre-training. In the pre-training experiment, no matter what the number of training epochs is, the models based on SciBERT had a lower perplexity than those based on BERT-base, indicating that the weight parameters of SciBERT are more suitable for social science papers. The SSCI-SciBERT-e2 and SSCI-SciBERT-e4 performed equally well on different tasks in the final evaluation, and it was impossible to determine with precision which of the two models performed better. However, the recognition results of them were better than those of the BERT-base training model. It shows a certain correlation between the perplexity and the performance of the model, but the correlation becomes more significant in the case of a larger difference in perplexity. However, when all the models have a relatively low perplexity value, perplexity cannot be used to predict model performance. Instead, it can only be used as a partial reference index.

In terms of classification, this research confirms that our proposed SsciBERT series models have an advantage in text classification tasks in the social sciences domain. It further verifies that the pre-trained models enhanced by domain data are able to perform intelligent information processing work better for the corresponding domain. Compared to the discipline classification results of the titles, the model classification results for the abstracts were more accurate, which suggests that the length of the input text intrinsically relates to the performance of the model on specific tasks. Longer input text may help the model to fully learn and extract text features. Therefore, future research should focus on using a larger-scale social science corpus to build the pre-trained models, e.g., using a full-text corpus of social science papers rather than restricting to bibliographic information such as abstracts. In addition, a purer domain pre-trained model may be constructed by training from scratch to better support social science text mining. In this research, the discipline of the journal that published the paper was identified as the discipline corresponding to the paper. Therefore, the discipline category of the abstract in the dataset is an annotation result based on a mapping relationship, which does not mean that the abstract of the paper is directly related to the discipline. Hence, the discipline classification results based on the pre-trained model may potentially be used to calculate the interdisciplinary nature of the literature and the journals, which is worthy of further research.

In terms of structure and functional recognition, the SsciBERT models presented in this paper improved the state-of-the-art classification of sentences according to the BPMRC paradigm. The Ssci-Scibert-e2 was the best-performing model. Among the models proposed in this paper, models further pre-trained from SciBERT had better performance than those from BERT-base, which indicates that the semantic computer and biomedical information in SciBERT is closer to social sciences text than general text. Also, the structural differences between the social sciences and general text in terms of abstract writing are also smaller. Within the BPMRC results, recognition performance between the method and result sentences is better than that of the background and purpose sentences. And the recognition performance of the background sentences is significantly lower than that of the method, result, and conclusion sentences, given the condition of a relatively balanced sentence sample size. This may be because the characteristics of the target sentences are not striking and are easier to get confused semantically with other types of sentences.

In order to further assess the performance of the SsciBERT, the software entity recognition models based on Bert-base-cased, Scibert-scivocab-cased, SSCI-BERT-e2, SSCI-SciBERT-e2, SSCI-BERT-e4 and SSCI-SciBERT-e4 were constructed for the academic full-text of *Scientometrics* that serves as a micro-language unit in natural language processing. In order to further validate the performance of the SsciBERT on domain-oriented social science data, this study manually constructed the largest corpora annotated with software entities in the field of informetrics by observing the distribution of software entities in the academic full-text of *Scientometrics*. The corpus annotated with software entities covers a long span of time and possesses remarkable attributes of the subject domain. In order to verify the performance of the SsciBERT comprehensively and concretely, the precision, recall and $F$1 score are used to evaluate the performance of the constructed software entity recognition model. The SSCI-BERT-e2 outperforms both the Bert-base-cased and the Scibert-scivocab-cased models in precision, recall, and $F$1 score, while its $F$1 score is also higher, respectively, going up by 1.53% and 0.79%. Compared with benchmark models such as Bert-base-cased and Scibert-scivocab-cased, the precision, recall, and $F$1 score of SSCI-BERT-e2, SSCI-SciBERT-e2, SSCI-BERT-e4, and SSCI- SciBERT-e4 exceeds 80.00%. Although the performance of the SsciBERT is better than that of the benchmark model, the acquired domain text knowledge of SsciBERT from titles and abstracts of academic literature is insufficient and incomplete compared with the academic full-text, which produces some negative effects on the performance of software entity recognition to a certain extent.

The pre-trained language model of social science academic text we proposed in this paper can promote data mining and the research of scientific evaluation for the social science text. Although this study only assessed the performance of SsciBERT concerning three natural language processing tasks, i.e. text classification, abstract structures identification, and software entity recognition, it can also be applied to text mining tasks such as text clustering, similarity calculation, discipline identification, knowledge entity summarization, relation extraction, etc. The reason is that the structure of this newly- proposed model exhibits a fine performance that is almost the same as that of BERT, SciBERT and other benchmark models. Furthermore, since the model of this research takes the abstract of the literature in the field of social science as the corpus and is formed on the basis of the continuous training of the original model, it is more suitable for the academic text mining of social science literature. It is found that there are great differences between academic texts in the field of social sciences, general texts and academic texts in natural sciences. One great advantage of this model is that it can generate better word vectors for academic texts in the social sciences, such as social science academic literature, policy reports, and academic forum records. Meanwhile, it also plays an important role in the context of full-text measurement and evaluation of academic texts. An appropriate deep learning framework could be selected and this model is used to initialize the parameters of the language model.

Accordingly, the following suggestions are put forward regarding how to choose the correct parameters in the process of using the model. First, several pre-experiments should be carried out to determine the suitable parameters for the target task, in which training epoch and learning rate deserve more attention. When the amount of training data is sufficient, there is no need to set the training epoch too large. While a small training epoch may lead to insufficient model training, a large training epoch may cause the model to overfit the training data. Ensuring that the loss curve shows a gradual decreasing trend when training the model after each update of the parameters. When the loss curve no longer decreases significantly, the corresponding training epoch is the appropriate epoch. As for the learning rate, it should avoid being too small, and the warm-up learning strategy is a recommended option. Second, the batch size of training data should be set as large as possible to fully utilize GPU resources and thus speed up training. At the same time, a larger batch data size is more conducive to the model to optimize the parameters. Briefly, the parameters used in this study are recommended for the users' first experiment. Finally, it is necessary to use high-quality datasets to train the model. The effect of a high-quality dataset on improving the performance of the model is much greater when compared with the adjustment of other parameters.

# 6 Conclusions

This paper proposes a pre-trained language model called SsciBERT for parsing academic texts relating to the social sciences. A large number of abstracts from the social sciences literature were collected from the SSCI database. These were used to further pre-train the already pre-trained models, i.e., BERT-base and SciBERT. With perplexity as an assessment criterion, the experimental results show that the perplexity of the SsciBERT model based on SciBERT is lower than that of the benchmark model. This indicates that the SsciBERT model, to some extent, is able to represent the semantic characteristics of social science academic texts better. In the three tasks designed further to compare the classification, identification and recognition of the models, results show that the SsciBERT model is able to improve upon the benchmarks.

Admittedly, this study also has some shortcomings, which serve as future research directions. First, considering the difficulty of automatically obtaining the full text of academic literature, only literature abstracts for pre-training are selected. Therefore, it is expected that a SsciBERT model trained on the full text of the articles would achieve better performance than that presented in this paper. Further research is considered for achieving the full text of academic literature in the authoritative journal in a different discipline for pre-training. Second, there are no standard datasets for text mining in the field of social sciences. For this reason, a self-built abstract classification and recognition dataset and software-tagged dataset with which to conduct the present experiments were adopted. Establishing a high-quality standard dataset of academic texts would better verify the performance of the model and deserve to be considered in future research. Finally, as to the progress of cross-language pre-training research, it is possible to obtain the texts of English, Chinese, and other multi-lingual academic literature for cross-language pre-training model training to enhance the semantic representation of the model from the perspective of different languages.

Overall, this research makes up for the lack of language models in the field of social sciences. This paper supports further collation, excavation, and utilization of academic texts in social sciences. It is also of significance to understand the evolution of discipline research content, the discovery of cross-disciplines, and emerging research growth points.

## Acknowledgements

The authors acknowledge the National Natural Science Foundation of China (Grant Numbers: 71974094, 72004169) for financial support and the data annotation team of Nanjing Agricultural University and Nanjing University of Science and Technology. Thanks to the students and researchers for their help in the revision and polishing of the paper.

## Conflict of interest statement

The authors declared that they have no conflicts of interest to this work.

# References

Asada, M., Miwa, M., & Sasaki, Y. (2020). Using drug descriptions and molecular structures for drug–drug interaction extraction from literature. *Bioinformatics, 37*(12), 1739-1746. doi:10.1093/bioinformatics/btaa907

Beltagy, I., Lo, K., & Cohan, A. (2019). *SciBERT: A pretrained language model for scientific text.* Paper presented at the 2019 Conference on Empirical Methods in Natural Language Processing and the 9th International Joint Conference on Natural Language Processing (EMNLP-IJCNLP 2019), Hong Kong.

Bengio, Y., Ducharme, R., & Vincent, P. (2000). *A neural probabilistic language model.* Paper presented at the Neural Information Processing Systems 2000 (NIPS 2000), Denver, Colorado.

Bojanowski, P., Grave, E., Joulin, A., & Mikolov, T. (2017). Enriching word vectors with subword information. *Transactions of the Association for Computational Linguistics, 5*, 135-146. doi:10.1162/tacl_a_00051

Brack, A., D'Souza, J., Hoppe, A., Auer, S., & Ewerth, R. (2020). Domain-independent extraction of scientific concepts from research articles. In *Advances in Information Retrieval* (pp. 251-266). Cham: Springer International Publishing.

Cattan, A., Johnson, S., Weld, D., Dagan, I., Beltagy, I., Downey, D., & Hope, T. (2021). *SciCo: Hierarchical cross-document coreference for scientific concepts.* Paper presented at the 3rd Conference on Automated Knowledge Base Construction (AKBC 2021), Irvine.

Chalkidis, I., Fergadiotis, M., Malakasiotis, P., Aletras, N., & Androutsopoulos, I. (2020). *LEGAL-BERT: The muppets straight out of law school.* Paper presented at the The 2020 Conference on Empirical Methods in Natural Language Processing (EMNLP 2020), Online.

Chen, S. F., Beeferman, D., & Rosenfeld, R. (1998). Evaluation metrics for language models.Paper presented at the Workshop of DARPA Broadcast News Transcription and Understanding,2-8.

D'Souza, J., Auer, S., & Pedersen, T. (2021, aug). *SemEval-2021 Task 11: NLPContributionGraph - Structuring scholarly NLP contributions for a research knowledge graph.* Paper presented at the 15th International Workshop on Semantic Evaluation (SemEval-2021), Online.

D'Souza, J., Hoppe, A., Brack, A., Jaradeh, M. Y., Auer, S., & Ewerth, R. (2020, may). *The STEM-ECR dataset: Grounding scientific entity references in STEM scholarly content to authoritative encyclopedic and lexicographic sources.* Paper presented at the 12th Language Resources and Evaluation Conference (LREC 2020), Marseille.

Devlin, J., Chang, M.-W., Lee, K., & Toutanova, K. (2019). *BERT: Pre-training of deep bidirectional transformers for language understanding.* Paper presented at the 17th Annual Conference of the North American Chapter of the Association for Computational Linguistics: Human Language Technologies (NAACL-HLT 2019), Minneapolis, Minnesota.

Dong, Q., Wan, X., & Cao, Y. (2021, apr). *ParaSCI: A Large scientific paraphrase dataset for longer paraphrase generation.* Paper presented at the 16th Conference of the European Chapter of the Association for Computational Linguistics (EACL 2021), Online.

Ferreira, D., & Freitas, A. (2020, may). *Natural language premise selection: Finding supporting statements for mathematical text.* Paper presented at the 12th Language Resources and Evaluation Conference (LREC 2020), Marseille.

Friedrich, A., Adel, H., Tomazic, F., Hingerl, J., Benteau, R., Marusczyk, A., & Lange, L. (2020). *The SOFC-Exp corpus and neural approaches to information extraction in the materials science domain.* Paper presented at the 58th Annual Meeting of the Association for Computational Linguistics (ACL 2020), Online.

Graetz, N. (1982). *Teaching EFL students to extract structural information from abstracts*. Paper presented at the International Symposium on Language for Special Purposes, Eindhoven.

He, K., Zhang, X., Ren, S., & Sun, J. (2016). *Deep residual learning for image recognition.* Paper presented at the 2016 IEEE Conference on Computer Vision and Pattern Recognition (CVPR 2016), Las Vegas, Nevada.

Hebbar, S., & Xie, Y. (2021, 04/18). *CovidBERT-Biomedical Relation Extraction for Covid-19.* Paper presented at the Florida Artificial Intelligence Research Society Conference, North Miami Beach, Florida.

Huang, K.-H., Yang, M., & Peng, N. (2020). *Biomedical event extraction with hierarchical knowledge graphs.* Paper presented at the 2020 Conference on Empirical Methods in Natural Language Processing (EMNLP 2020), Online.

Kononova, O., He, T., Huo, H., Trewartha, A., Olivetti, E. A., & Ceder, G. (2021). Opportunities and challenges of text mining in materials research. *iScience, 24*(3), 102155. doi:10.1016/j.isci.2021.102155

Kotonya, N., & Toni, F. (2020). *Explainable automated fact-checking for public health claims.* Paper presented at the 2020 Conference on Empirical Methods in Natural Language Processing (EMNLP), Online.

Koutsikakis, J., Chalkidis, I., Malakasiotis, P., & Androutsopoulos, I. (2020). *GREEK-BERT: The Greeks visiting Sesame Street.* Paper presented at the 11th Hellenic Conference on Artificial Intelligence (SETN 2020), Athens.

Kuniyoshi, F., Makino, K., Ozawa, J., & Miwa, M. (2020). *Annotating and Extracting Synthesis Process of All-Solid-State Batteries from Scientific Literature.* Paper presented at the 12th Language Resources and Evaluation Conference (LREC 2020), Marseille.

Lauscher, A., Ko, B., Kuehl, B., Johnson, S., Jurgens, D., Cohan, A., & Lo, K. (2021). MultiCite: Modeling realistic citations requires moving beyond the single-sentence single-label setting. *arXiv preprint arXiv:2107.00414*.

Lee, J., Yoon, W., Kim, S., Kim, D., Kim, S., So, C. H., & Kang, J. (2019). BioBERT: a pre-trained biomedical language representation model for biomedical text mining. *Bioinformatics, 36*(4), 1234-1240. doi:10.1093/bioinformatics/btz682

Medić, Z., & Šnajder, J. (2020). A survey of citation recommendation tasks and methods. *Journal of computing and information technology, 28*(3), 183-205. doi:10.20532/cit.2020.1005160

Mikolov, T., Chen, K., Corrado, G., & Dean, J. (2013). Efficient estimation of word representations in vector space. *arXiv preprint arXiv:1301.3781*.

Muraina, I. (2022). *IDEAL DATASET SPLITTING RATIOS IN MACHINE LEARNING ALGORITHMS: GENERAL CONCERNS FOR DATA SCIENTISTS AND DATA ANALYSTS.*

Murty, S., Koh, P. W., & Liang, P. (2020, jul). *ExpBERT: Representation Engineering with Natural Language Explanations*, Online.

Nicholson, J. M., Mordaunt, M., Lopez, P., Uppala, A., Rosati, D., Rodrigues, N. P., . . . Rife, S. C. (2021). Scite: A smart citation index that displays the context of citations and classifies their intent using deep learning. *Quantitative Science Studies, 2*(3), 882-898. doi:10.1162/qss_a_00146

Park, S., & Caragea, C. (2020). *Scientific keyphrase identification and classification by pre-trained language models intermediate task transfer learning.* Paper presented at the 28th International Conference on Computational Linguistics (COLING'2020), Barcelona (Online).

Pennington, J., Socher, R., & Manning, C. (2014). *GloVe: Global Vectors for Word Representation.* Paper presented at the 2014 Conference on Empirical Methods in Natural Language Processing (EMNLP 2014), Doha.

Peters, M. E., Neumann, M., Iyyer, M., Gardner, M., Clark, C., Lee, K., & Zettlemoyer, L. (2018). *Deep contextualized word representations*, New Orleans, Louisiana.

Radford, A., Narasimhan, K., Salimans, T., & Sutskever, I. (2018). Improving language understanding by generative pre-training. Retrieved from https://www.cs.ubc.ca/~amuham01/LING530/papers/radford2018improving.pdf

Rasmy, L., Xiang, Y., Xie, Z., Tao, C., & Zhi, D. (2021). Med-BERT: pretrained contextualized embeddings on large-scale structured electronic health records for disease prediction. *NPJ Digital Medicine, 4*(1), 86. doi:10.1038/s41746-021-00455-y

Simonyan, K., & Zisserman, A. (2014). Very deep convolutional networks for large-scale image recognition. doi:10.48550/arXiv.1409.1556

Sollaci, L. B., & Pereira, M. G. (2004). The introduction, methods, results, and discussion (IMRAD) structure: a fifty-year survey. *Journal of the Medical Library Association, 92*(3), 364-367.

Swales, J. (1990). *Genre analysis: English in academic and research settings*. Cambridge: Cambridge University Press.

van Dongen, T., Maillette de Buy Wenniger, G., & Schomaker, L. (2020, nov). *SChuBERT: Scholarly document chunks with BERT-encoding boost citation count prediction.* Paper presented at the 1st Workshop on Scholarly Document Processing (SDP 2020), Online.

Vaswani, A., Shazeer, N., Parmar, N., Uszkoreit, J., Jones, L., Gomez, A. N., . . . Polosukhin, I. (2017). *Attention is all you need.* Paper presented at the The 31 Annual Conference on Neural Information Processing Systems (NIPS), Long Beach, California.

Viswanathan, V., Neubig, G., & Liu, P. (2021, aug). *CitationIE: Leveraging the citation graph for scientific information extraction.* Paper presented at the 59th Annual Meeting of the Association for Computational Linguistics and the 11th International Joint Conference on Natural Language Processing (ACL-IJCNLP 2021)), Online.

Wolf, T., Debut, L., Sanh, V., Chaumond, J., Delangue, C., Moi, A., ... & Rush, A. M. (2019). Huggingface's transformers: State-of-the-art natural language processing. arXiv preprint arXiv:1910.03771.

Wright, D., & Augenstein, I. (2021). *CiteWorth: Cite-worthiness detection for improved scientific document understanding.* Paper presented at the The Joint Conference of the 59th Annual Meeting of the Association for Computational Linguistics and the 11th International Joint Conference on Natural Language Processing (ACL-IJCNLP 2021), Online.

Yang, Y., Siy UY, M. C., & Huang, A. (2020). FinBERT: A pretrained language model for financial communications. doi:10.48550/arXiv.2006.08097